# Improving Supervised Machine Learning Performance in Optical Quality Control via Generative AI for Dataset Expansion

Dennis Sprute*, Hanna Senke, Holger Flatt

*Fraunhofer IOSB, Industrial Automation Branch (IOSB-INA), 32657 Lemgo, Germany*

* Corresponding author. Tel.: +49 5261 94290 63. *E-mail address:* dennis.sprute@iosb-ina.fraunhofer.de

**Abstract**

Supervised machine learning algorithms play a crucial role in optical quality control within industrial production. These approaches require representative datasets for effective model training. However, while non-defective components are frequent, defective parts are rare in production, resulting in highly imbalanced datasets that adversely impact model performance. Existing strategies to address this challenge, such as specialized loss functions or traditional data augmentation techniques, have limitations, including the need for careful hyperparameter tuning or the alteration of only simple image features. Therefore, this work explores the potential of generative artificial intelligence (GenAI) as an alternative method for expanding limited datasets and enhancing supervised machine learning performance. Specifically, we investigate Stable Diffusion and CycleGAN as image generation models, focusing on the segmentation of combine harvester components in thermal images for subsequent defect detection. Our results demonstrate that dataset expansion using Stable Diffusion yields the most significant improvement, enhancing segmentation performance by 4.6 %, resulting in a Mean Intersection over Union (Mean IoU) of 84.6 %.

**1. Introduction**

Quality control is a critical step in machinery production, where identifying defective components is essential. For this purpose, supervised machine learning algorithms are often employed to detect and classify components in images [1]. These approaches require large amounts of representative annotated data for model training to achieve a strong performance. However, while non-defective components are common, defective parts are rare in production, leading to highly imbalanced datasets. This imbalance poses a challenge for supervised machine learning algorithms and negatively impacts the model performance. Acquiring a sufficient number of defective components is often time-consuming, expensive, or even impossible. Various methods have been proposed to address dataset imbalance, such as unsupervised learning approaches like anomaly detection [2], the application of specialized loss functions during neural network training, such as focal loss [3], and traditional data augmentation techniques [4]. However, anomaly detection algorithms struggle to differentiate between multiple defective classes, applying focal loss requires careful hyperparameter tuning, and traditional data augmentation only modifies simple image features, such as color and geometry. Therefore, we propose an alternative approach to tackle this data imbalance problem by exploring the potential of generative artificial intelligence (GenAI) for dataset expansion. The main research question of this work is: *Can GenAI methods be* *employed for artificial dataset expansion to improve the performance of supervised machine learning models?*

In order to address this question, the use case of semantic segmentation of combine harvester components in thermal images is considered [5]. The generation of heat by these components in the presence of defects requires their localization in the image prior to the application of a temperature threshold for defect detection. Exemplary images are shown in Fig. 1. This use case is well-suited to answer the research question, as defective parts are rare, and acquiring additional images of defective machines is extremely time-consuming and costly. The main idea is to transform images of non-defective machines into images of defective machines using GenAI methods, thereby balancing the dataset and ultimately improving semantic segmentation performance.

## 2. Related work

GenAI focuses on the artificial generation of data, including text, images, and videos, through the use of generative models. A notable category within this field are models for image generation, such as Stable Diffusion [6] and DALL-E [7], which produce images based on textual input. The recent progress and impressive results have sparked interest in expanding image datasets artifically. For example, Zhang et al. proposed a Guided Imagination Framework that automatically generates new labeled data samples for small datasets [8]. Azizi et al. augmented the ImageNet dataset using a generative diffusion model to enhance classification performance [9]. Additionally, a Generative Adversarial Network (GAN) has been employed in an active learning approach to create images and improve classification outcomes [10]. Such general approaches have also been transferred to more specific application domains that face imbalanced data challenges. For instance, CycleGAN has been utilized for data augmentation in the medical field, transforming contrast-enhanced CT images into non-contrast images to improve segmentation performance [11]. Similarly, Yigit and Can leveraged Stable Diffusion to generate a high-quality infrared dataset for hazardous situations in autonomous driving [12].

These studies demonstrate that GenAI methods have proven beneficial for dataset expansion, particularly in domains struggling with imbalanced data, such as medical image processing and autonomous driving. However, to the best of our knowledge, there has been no systematic investigation into the impact of artificially generated images on the performance of supervised machine learning methods within the context of industrial production, specifically regarding thermal images and optical quality control.

## 3. Materials and methods

An overview of the proposed methodology is illustrated in Fig. 2. The starting point is a highly imbalanced dataset of real images. The primary objective of this work is to utilize images from the underrepresented class(es) and transform them into images of the overrepresented class(es) through artificial image generation. The outcome is an expanded dataset with a more balanced class distribution, which is expected to enhance the performance of supervised machine learning methods.

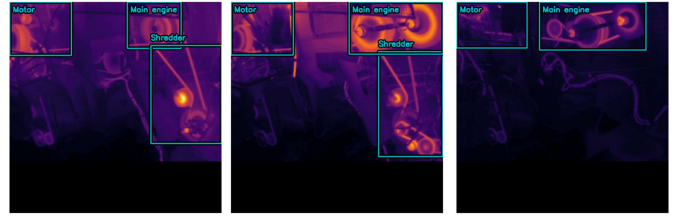

Fig. 1 Exemplary thermal images of different combine harvester variants

### 3.1. Real image dataset

The baseline dataset comprises raw temperature data and corresponding color-mapped thermal images of combine harvesters [5]. The data collection process spanned 49 production days over a period of 7 months, yielding a total of 1,200 images. Among these, 253 images contained defective components, resulting in an imbalanced dataset. The images feature side views of combine harvesters and their components, which include the motor, main engine, and shredder. Different variants of these components can be combined in various ways: there is one motor variant, five distinct main engines labeled A to E, and five types of shredders labeled 1 to 5, where the number 1 indicates machines without a shredder. In total, there are 19 different machine variants, three of which are depicted in Fig. 1. Among these, five machine variants include defective machines, along with four different types of defective components and their combinations. This results in nine distinct variations of defective machines. Each image is annotated with a segmentation mask of the components.

### 3.2. Image generation models

To expand the real dataset with artificial data, two image generation models are considered.

#### 3.2.1. Stable Diffusion
Stable Diffusion is a diffusion model that generates images in latent space [6]. It can create artificial images from text prompts or modify existing images based on text guidance. The core components of Stable Diffusion include a text encoder, a diffusion model, and an image encoder and decoder.

However, the current Stable Diffusion models are not suitable for thermal images of combine harvester components, as they are primarily trained on visible image data. Additionally, the interior of a combine harvester does not represent common objects found in the training dataset. Therefore, fine-tuning of the model is necessary. During the fine-tuning process, the text encoder and image encoder are kept frozen, while only the diffusion model is trained. To fine-tune Stable Diffusion, images with corresponding text prompts are required. For defective machines, two prompt variants are tested:

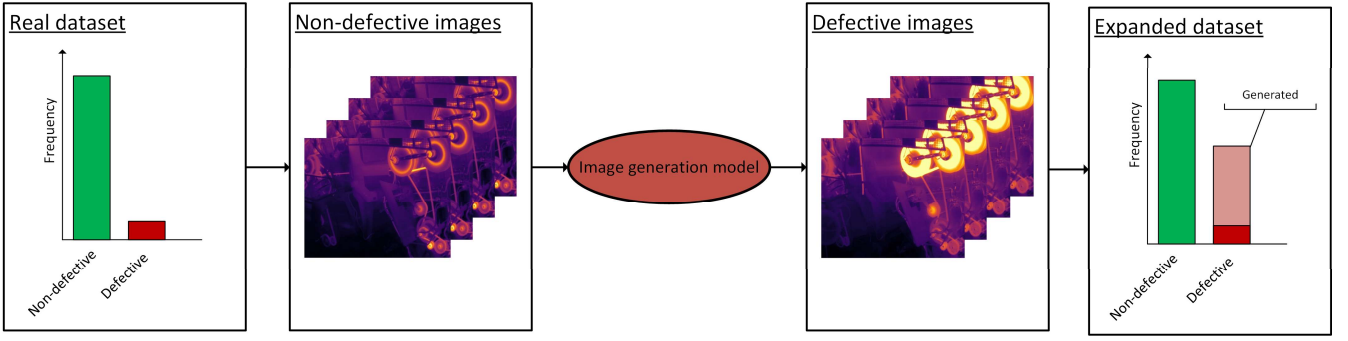

Fig. 2 Overview of the proposed approach expanding an image dataset artificially via an image generation model

- Variant 1: "a {image-type} image of a {variant} ch machine with a defective {component-name}"
- Variant 2: "a {image-type} image of a defective {variant} ch machine"

The parameter {variant} specifies one of the combine harvester variants described in Section 3.1, while {image-type} can be either "rgb" or "grayscale". In the first prompt, the defective component is named according to one of the three components or a combination of them. In the second prompt, the component is not specified.

For the images of non-defective combine harvesters, only one prompt variant is used:

- "a {image-type} image of an intact {variant} ch machine"

The term "combine harvester" is known to Stable Diffusion prior to fine-tuning, but it describes the exterior of the machine rather than the interior view. Since the exterior view has no correlation with the interior view, the expression "ch" is introduced in these prompts.

### 3.2.2. CycleGAN

CycleGAN is a Generative Adversarial Network (GAN) designed for unpaired image-to-image translation, consisting of two GAN models [13]. Each GAN comprises a generator for image conversion and a discriminator to distinguish between real and generated images. Similar to Stable Diffusion, CycleGAN requires fine-tuning for the specific use case. For this purpose, CycleGAN needs an unpaired image dataset for training, specifying the input and target domains, i.e. images of non-defective and defective machines.

### 3.3. Training of image generation models

To create a dataset for model training, up to 8 images are selected per defective combine harvester, resulting in a total of 40 images. This selection is performed manually to ensure a representative collection of defective machines with various defective components. From the images of non-defective machines, up to 5 images per variant are also chosen manually, covering 18 of the 19 combine harvester variants, which leads to a total of 129 images in the training dataset. Each image in the dataset is labeled with the textual prompts described in Section 3.2.1, and the dataset contains each image in both RGB and grayscale formats.

Both image generation models are then trained on the training datasets using standard hyperparameters as implemented by the software libraries detailed in Section 4.2. Due to the limited number of defective images in the training dataset compared to non-defective images, an additional two-stage approach is employed for the training of Stable Diffusion. In this approach, the model is initially trained on the entire dataset and then further fine-tuned specifically on the defective images. The rationale is to first train on all machine variants and subsequently focus on generating defective machines. It is necessary to train on the entire dataset first in this two-stage process to ensure that all variants are present in the dataset.

### 3.4. Image generation

For image generation, a dataset of non-defective combine harvester images is required, which will be transformed into images of defective combine harvesters. For each machine variant, up to 5 images of non-defective machines, not included in the training dataset, are selected. In total, this dataset comprises 80 images, each available in both RGB and grayscale formats.

Using Stable Diffusion, all images are generated through the image-to-image generation process to retain the component locations of the real dataset. Various hyperparameters can be adjusted to fine-tune the generation process, summarized in Table 1. The first hyperparameter is the prompt variant described in Section 3.2.1. For Variant 1, a separate dataset is created for each of the three components. The image type and fine-tuning hyperparameters pertain to the training process of the image generation model, allowing for the selection between models trained on RGB and grayscale images and those that include additional fine-tuning on defective images. Moreover, the unconditional guidance scale (UGS) can be modified. A higher UGS value results in the generated image adhering more closely to the provided text input. Two different UGS values are investigated in this work, i.e. 7.5 and 13.0. The input image strength determines how much influence the input image has on the generated output. A lower value makes the output more similar to the input image. The standard value for input image strength is set at 0.5, but in this work, values of 0.3, 0.5, and 0.7 are tested. Additionally, the number of images generated per prompt are set to 2, and the number of diffusion model steps is set to the standard value of 50.

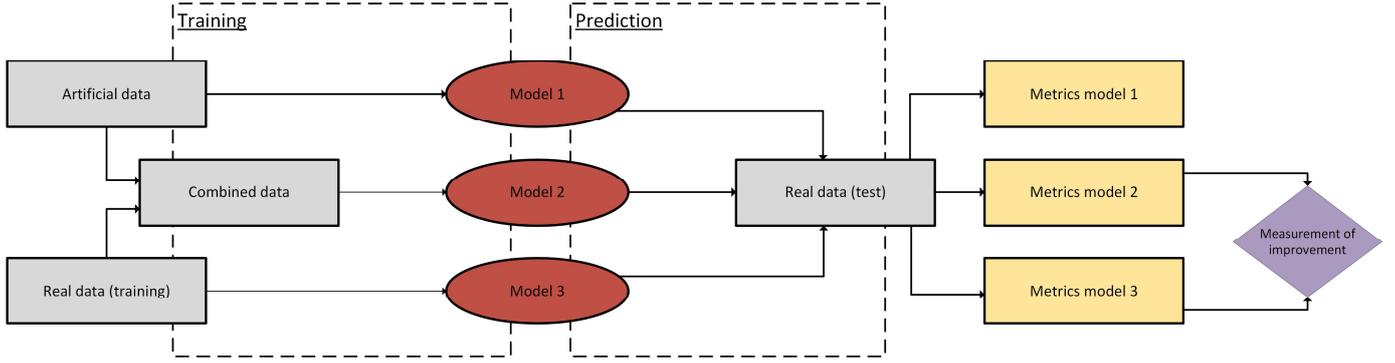

Fig. 3 Overview of the experimental design

Table 1 Hyperparameters for Stable Diffusion image generation process

| Hyperparameter | Options |
|---|---|
| Prompt variant | {Variant 1; Variant 2} |
| Image type | {RGB; Gray} |
| Fine tuning | {True; False} |
| Unconditional guidance scale (UGS) | {7.5; 13.0} |
| Image strength | {0.3; 0.5; 0.7} |
| Number of output images per input image | {2} |
| Number of steps | {50} |

For CycleGAN, only the image type and the number of images per prompt can be set, as summarized in Table 2. These parameters are consistent with those used in Stable Diffusion.

Table 2 Hyperparameters for CycleGAN image generation process

| Hyperparameter | Options |
|---|---|
| Image type | {RGB; Gray} |
| Number of output images per input image | {2} |

## 4. Experimental evaluation

The main objective of the experimental evaluation is to test the following hypothesis: *Expanding an image dataset using GenAI methods can improve the performance of supervised machine learning.*

### 4.1. Setup and design

To investigate this hypothesis, we designed the experiment as illustrated in Fig. 3. First, a model (*Model 1*) is trained in a supervised manner on the artificial data generated by the image generation models. This model is then used to make predictions on a test subset of the real data and is evaluated. This initial procedure serves as a quality check to determine if the model trained on artificial data can be effectively applied to real data, allowing for the selection of the best hyperparameters for image generation. Following this, a second model (*Model 2*) is trained on a combined dataset, consisting of a training subset of the real data and the artificially generated data. Model 2 is then also evaluated on the test subset of the real data. The same procedure applies to a third model (*Model 3*), which is solely trained on the real data. The final step involves comparing the performance of *Model 2* and *Model 3* to test our hypothesis.

All supervised machine learning models trained are SegFormer [14] models for semantic segmentation, as this architecture was identified as the best-performing model for this use case [5]. The model utilizes an MiT-B0 backbone with pre-trained weights on the ImageNet dataset. The evaluation metrics calculated include the Mean Intersection over Union (Mean IoU), a common metric for semantic segmentation.

### 4.2. Hardware and software

Experiments were conducted on a desktop PC equipped with a 12th Gen Intel Core i7 processor running at a base speed of 3.60 GHz, 64 GB of RAM, and an NVIDIA GeForce RTX 3060 graphics card with 12 GB of VRAM. The implementation is based on Keras (v3.0.5) and TensorFlow (2.16.1). As the Stable Diffusion model provided by the Keras API did not support image-to-image translation at the time of implementation, an alternative image-to-image implementation was chosen [15]. For CycleGAN, the implementation provided by Keras was utilized [16].

### 4.3. Results

The results of the first part of the experiment, which involved training models (*Model 1*) exclusively on the artificial data, are summarized in Table 3. This table lists the segmentation performance for the two best-performing hyperparameter combinations per prompt variant for Stable Diffusion, as well as the results based on the image type for CycleGAN. The findings indicate that RGB is the preferred image type for both image generation models. For Stable Diffusion, a UGS of 7.5 and an image strength of 0.3 lead to the best results across most of the prompt variants. Furthermore, additional fine-tuning using only defective images has only a minor impact on performance. Similarly, the variations in prompt types for Stable Diffusion demonstrate a limited effect on the outcomes.

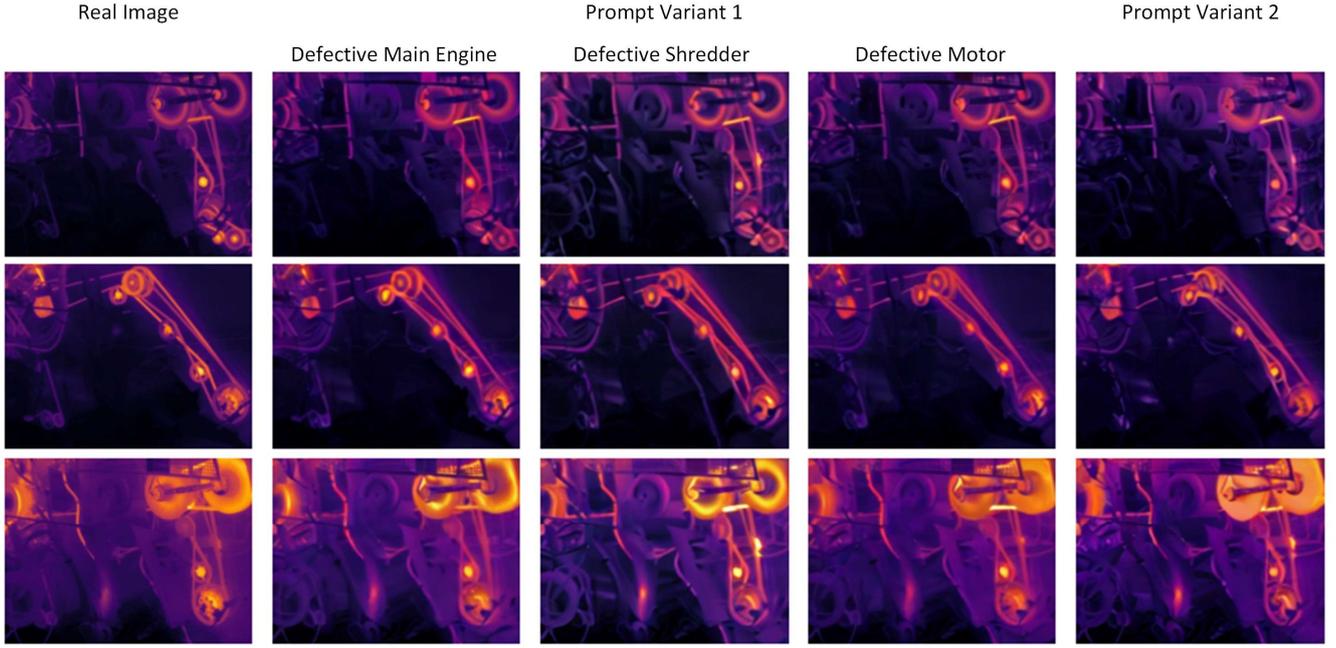

Fig. 4 Examples of generated images using Stable Diffusion and different prompt variants

Table 3 SegFormer performance: Training on artificial datasets and prediction on test subset of real dataset

| Method | Prompt | Image type | Fine tuning | UGS | Image strength | Mean IoU |
|---|---|---|---|---|---|---|
| Stable Diffusion | V1: Main engine | RGB | False | 7.5 | 0.3 | 0.649 |
| Stable Diffusion | V1: Main engine | RGB | False | 13.0 | 0.3 | 0.640 |
| Stable Diffusion | V1: Shredder | RGB | False | 7.5 | 0.3 | 0.654 |
| Stable Diffusion | V1: Shredder | RGB | True | 7.5 | 0.3 | 0.653 |
| Stable Diffusion | V1: Motor | RGB | False | 7.5 | 0.3 | 0.654 |
| Stable Diffusion | V1: Motor | RGB | True | 7.5 | 0.3 | 0.639 |
| Stable Diffusion | V2 | RGB | True | 7.5 | 0.3 | 0.654 |
| Stable Diffusion | V2 | RGB | False | 13.0 | 0.3 | 0.632 |
| CycleGAN | --- | RGB | --- | --- | --- | 0.652 |
| CycleGAN | --- | Gray | --- | --- | --- | 0.553 |

The results of the second evaluation step, which compares a model trained on a combined dataset (*Model 2*) with a model trained exclusively on real data (*Model 3*), are presented in Table 4. For Model 2, the best generated dataset identified in Table 3 is combined with the real data. The baseline approach (*Model 3*) achieves a Mean IoU of 0.8. This performance is improved by 4.6 % when augmenting the dataset of real images with artificially generated images from Stable Diffusion and by 3.1 % when using CycleGAN. Thus, we conclude that these results support our hypothesis that expanding an image dataset using GenAI methods can enhance the performance of supervised machine learning.

Qualitative results of the image generation are illustrated in Fig. 4 for Stable Diffusion and Fig. 5 for CycleGAN.

Table 4 SegFormer performance: Training on given datasets and prediction on test subset of real dataset

| Dataset | Mean IoU |
|---|---|
| Real data | 0.800 |
| Real data + Stable diffusion | 0.846 (+4.6 %) |
| Real data + CycleGAN | 0.831 (+3.1 %) |

*4.4. Discussion*

The qualitative results indicate that there are only minor differences between the generated and real images for both image generation models. The components in the generated images appear brighter, suggesting higher temperatures, which serve as an indicator of defective components. This implies that defective images can effectively be generated from non-defective ones. However, when examining prompt variant 1 for Stable Diffusion, it becomes evident that defects are occasionally generated on incorrect components. For instance, when prompted to generate a defective shredder, there is also an unintended increase in brightness at the main engine. Additionally, generating defective motors proves challenging, as there is only a single variant in the real dataset exhibiting such a defect, resulting in the image generation model's inability to learn this adequately. Similarly, CycleGAN generates images of defective components by increasing their brightness. Due to the absence of prompt variants for specific components, the model tends to produce images where all components appear defective.

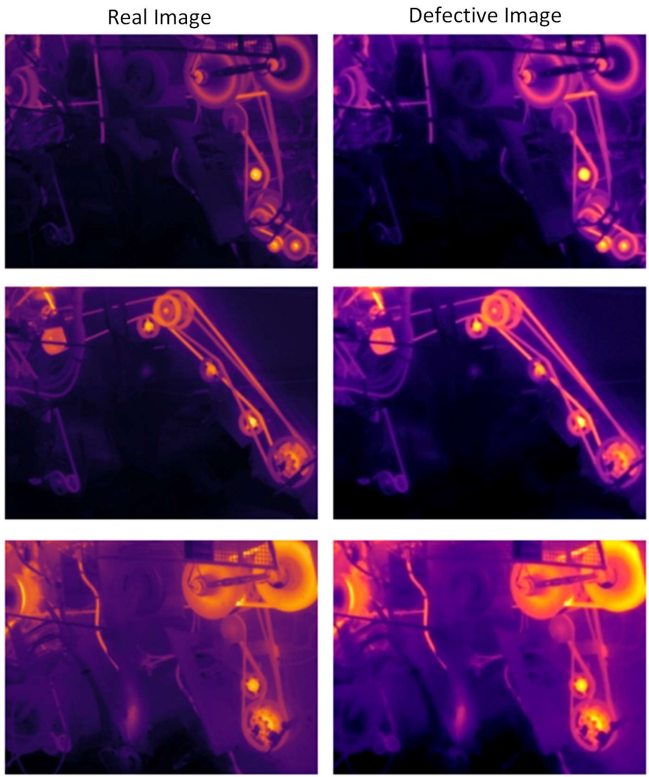

Fig. 5 Examples of generated images using CycleGAN

## 5. Conclusions

In this study, we investigated whether the performance of supervised machine learning models can be enhanced through artificial dataset expansion using GenAI methods. Our experiments demonstrated that Stable Diffusion is particularly effective for expanding a dataset with artificial data, resulting in a 4.6 % improvement in the performance of a SegFormer semantic segmentation model, achieving a Mean IoU of 0.846. This finding is especially relevant for optical quality control applications in industrial production, where images of defective components are often scarce. Thus, GenAI can be utilized to transform images of non-defective components into images of defective ones, effectively balancing the dataset without the need for time-consuming and costly data acquisition, thereby boosting the performance of supervised machine learning.

While this work focused on the specific use case of semantic segmentation in thermal data, further research should examine the generalizability of these findings to other supervised machine learning tasks, such as image classification or object detection. Additionally, future work should address the limitations of the current approach, particularly in improving the generation of images featuring single defective components. To this end, acquiring additional real images of defective components would be beneficial, thereby enriching the dataset for training the image generation models with more variants of defective combine harvesters.

## Acknowledgements

This research work is based on "Datenfabrik.NRW", a flagship project by "KI.NRW", funded by the Ministry for Economics, Innovation, Digitisation and Energy of the State of North Rhine-Westphalia (MWIDE). We like to thank our project partner CLAAS for supporting the image data acquisition.